\documentclass{article}
\usepackage{spconf}

\usepackage{enumitem}
\setlist{nosep, leftmargin=14pt}
\usepackage{afterpage}

\usepackage{mwe} 
\usepackage{amssymb,amsthm,enumitem}
\usepackage{multirow}
\usepackage{layouts}
\usepackage[ruled,linesnumbered,noend]{algorithm2e}
\usepackage{enumerate}

\usepackage{hyperref}
\usepackage{xcolor}
\hypersetup{
    colorlinks,
    linkcolor={blue!50!black},
    citecolor={blue!50!black},
    urlcolor={blue!50!black}
}
\usepackage{color}

\usepackage{bbm}
\usepackage{amsmath}
\usepackage{adjustbox}

\newcommand\blfootnote[1]{%
  \begingroup
  \renewcommand\thefootnote{}\footnote{#1}%
  \addtocounter{footnote}{-1}%
  \endgroup
}
\newcommand\blankpage{
    \null
    \thispagestyle{empty}
    \addtocounter{page}{-1}
    \newpage
    }

\title{BEL: A Bag Embedding Loss for Transformer enhances Multiple Instance Whole Slide Image Classification}

\name{Daniel Sens$^{1,\ast}$
    \quad Ario Sadafi$^{1,2,\ast}$
    \quad Francesco Paolo Casale$^{1,2}$
    \quad Nassir Navab$^{2,3}$
    \quad Carsten Marr$^{1}$}
\address{$^{1}$Helmholtz Munich, Neuherberg, Germany \\
        $^{2}$Faculty of Informatics, Technical University of Munich, Germany \\
        $^{3}$Computer Aided Medical Procedures, Johns Hopkins University, USA \\
        }
        
\begin{document}
%
\onecolumn

\hspace{0pt}
\vfill

{\Huge IEEE Copyright Notice}
\\[0.7in]
{\large © 2023 IEEE. Personal use of this material is permitted. Permission from IEEE must be obtained for all other uses, in any current or future media, including reprinting/ republishing this material for advertising or promotional purposes, creating new collective works, for resale or redistribution to servers or lists, or reuse of any copyrighted component of this work in other works.}
\\[.3in]
\textbf{\large
Pre-print of article that will appear at the 2023 IEEE International Symposium on Biomedical Imaging (ISBI 2023), April 18-21 2023}
\hspace{0pt}
\vfill

\blankpage{}
\twocolumn
\maketitle
\blfootnote{$^{\ast}$shared first authorship}
\begin{abstract}
 Multiple Instance Learning (MIL) has become the predominant approach for classification tasks on gigapixel histopathology whole slide images (WSIs). Within the MIL framework, single WSIs (bags) are decomposed into patches (instances), with only WSI-level annotation available. Recent MIL approaches produce highly informative bag level representations by utilizing the transformer architecture's ability to model the dependencies between instances. However, when applied to high magnification datasets, problems emerge due to the large number of instances and the weak supervisory learning signal. To address this problem, we propose to additionally train transformers with a novel Bag Embedding Loss (BEL). BEL forces the model to learn a discriminative bag-level representation by minimizing the distance between bag embeddings of the same class and maximizing the distance between different classes. We evaluate BEL with the Transformer architecture TransMIL on two publicly available histopathology datasets, BRACS and CAMELYON17. We show that with BEL, TransMIL outperforms the baseline models on both datasets, thus contributing to the clinically highly relevant AI-based tumor classification of histological patient material.
\end{abstract}
\begin{keywords}
Transformer, Multiple Instance Learning, Whole Slide Imaging, Bag-level Representation Learning, Computational Pathology
\end{keywords}
\section{Introduction}
\begin{figure*}[h!]
    \centering
    \includegraphics[width=0.72\textwidth]{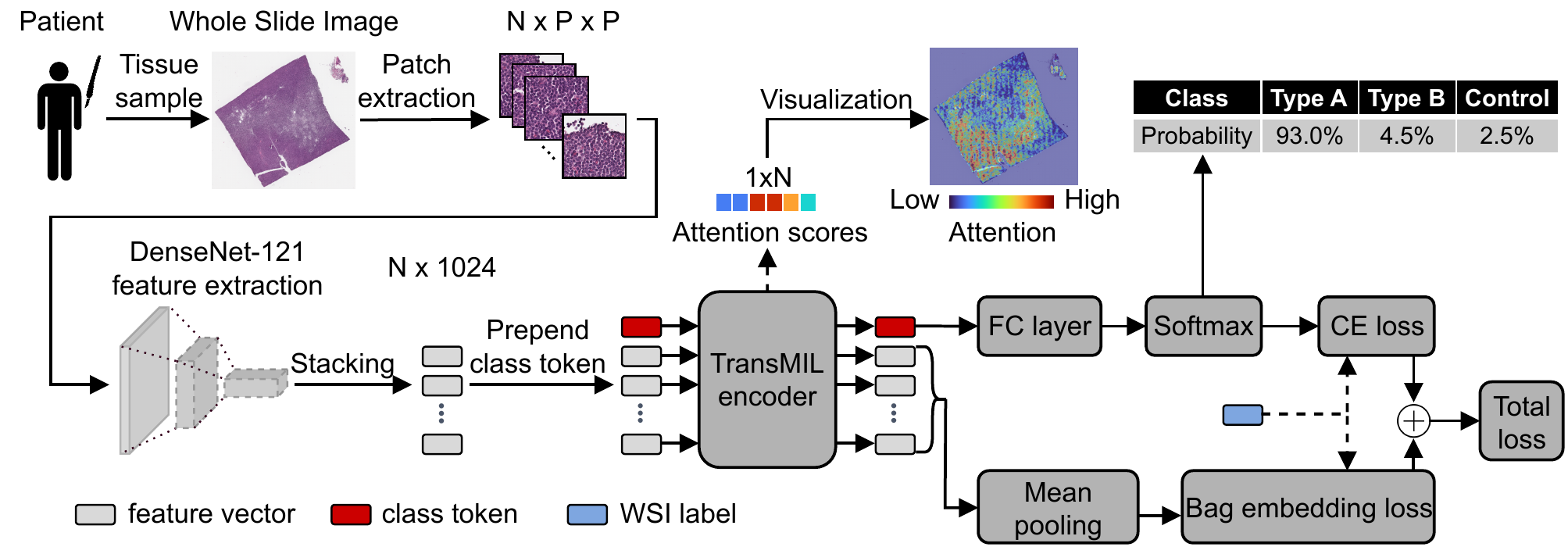}
    \caption{Overview of the proposed pipeline. Tissue samples from a patient are digitized to obtain a whole slide image (WSI). $N$ Patches of size $P \times P$ are extracted and sequentially processed by a pretrained DenseNet-121 for feature extraction. The resulting $N$ feature vectors are stacked and a class token is prepended to the sequence. This sequence is passed to the TransMIL encoder that projects each element to a latent representation. We obtain class probabilities by appending a fully connected (FC) layer on the output class token and a softmax thereafter. Additionally, we extract attention scores to visualize important regions of the WSI. During training of the TransMIL, we combine cross entropy (CE) loss with the proposed bag embedding loss (BEL) to enhance performance.}
    \label{con_fig}
\end{figure*}
Analysis of histopathological images has become indispensable in modern medicine, especially in cancer, where it is key for the correct diagnosis and treatment of the patients. With the advent of digital slide scanners, histology slides are digitized into gigapixel whole slide images (WSIs). Due to their size, WSIs are a challenge for computational analysis, and thus usually divided into patches to facilitate further processing. However, patch-wise annotations from expert pathologists are expensive, and often only a single label per WSI, is available. Additionally, relevant information is not distributed evenly among all parts of the WSI. Computational approaches are thus commonly formulated as a multiple instance learning (MIL) task, where a WSI is regarded as a bag, the extracted patches are regarded as the instances and only bag-level labels are available. 
Recent MIL approaches for pathology tend to adopt the following two-step procedure: (i) extract instance features from patches and (ii) aggregate the instance features to generate bag level representations. The first attempts relied on fixed aggregation functions such as mean pooling and max pooling with limited flexibility~\cite{pooling1,pooling2}. To address this, Ilse et al.~\cite{ABMIL} proposed attention pooling, which computes the bag representation based on a weighted average of the instance features, where the weights are learned by an attention mechanism during training. The clustering-constrained attention MIL (CLAM) proposed by Lu et al.~\cite{CLAM} introduces the concept of multiple attention branches, each with an additional loss computed based on pseudo-labels of the instance. However, the weight of an instance is still inferred solely from the instance itself and hence does not take into account the dependency between the instances, which is naturally done by the pathologists when making clinical decisions. 
To remedy this, Li et al.~\cite{DSMIL} proposed the Dual-Stream MIL (DSMIL), which uses non-local attention by calculating the similarity between the highest-score instance and others. Despite currently being one of the most advanced MIL approaches, this still does not take into account the dependencies between all instances in a bag. Myronenko et al.~\cite{ADMIL} thus proposed to embed transformer encoder blocks into a classification CNN. The inherent self-attention layers allow the transformer to update each element of a sequence by aggregating all elements in the sequence simultaneously. Because of their permutation invariance, however, self-attention layers alone are not sufficient to capture the spatial structure of patches in a WSI. Motivated by this issue Li et al.~\cite{DTMIL} proposed a fully end-to-end trainable encoder-decoder based transformer architecture, embedding a 2D positional encoding into the instance features. 
However, all of the above mentioned methods have so far not been applied to WSIs at maximum magnification, because the resulting number of instances would be too large to be processed by traditional self-attention layers due to their computational complexity. A possible solution is to work at a lower magnification which reduces the amount of patches. This, in turn, thwarts the explainability of transformer models, because the respective patches often contain too many cells or appear too blurry to identify meaningful morphological details at the single cell level. Furthermore, the absence of an additional, more sophisticated supervised learning signal tempts the transformer to overfit, as it heavily relies on large data sets that are often not available for WSI classification task.
In this work we address the two problems stated above by applying the transformer encoder based TransMIL model ~\cite{transmil} to WSIs at the highest, $40\times$ magnification. TransMIL makes use of the recently proposed Neystr\"om method~\cite{nystroemformer} to approximate self-attention, which reduces the computational complexity and is thus able to process bags with a huge number of instances. Specifically, we propose a pipeline where digitized tissue WSIs are segmented from the background and patches are extracted at the highest magnification. We then extract instance features from each patch using a pretrained DenseNet-121~\cite{densenet} and train TransMIL to classify the WSIs based on the extracted features in a MIL fashion (Fig.~\ref{con_fig}). Furthermore, we propose a novel Bag Embedding Loss (BEL) which is added to the classification loss during training, thereby forcing TransMIL to learn discriminative bag level representations. Extensive evaluations on the BRACS and CAMELYON17 dataset show that when trained with BEL, TransMIL is able to outperform the baseline models. Finally, we analyze the impact of BEL on the attention distribution of TransMIL.

\section{Methods}
Formally, the TransMIL $f_{\mathrm{TMIL}}$ takes a set of $N$ instance feature vectors of size $H$,
$X=\{x_1,\ldots,x_N\},\,x_i \in \mathbb{R}^{H}$ obtained from a WSI and outputs the predicted class probabilities $p \in \mathbb{P}$, a bag embedding $b\in \mathbb{R}^{H}$, and an attention vector $\alpha \in [0,1]^N$ that denotes the contribution of each instance to the final decision: 
\begin{equation}\label{eq:TMIL}
    p, b, \alpha  = f_{\mathrm{TMIL}}(X; \theta)\text{.}
\end{equation}
Here $\theta$ are the model parameters learned during the training using the total loss
\begin{equation}\label{eq:tloss}
    \mathcal{L}_{\mathrm{TOTAL}}(\theta) = \mathcal{L}_{\mathrm{CE}}(p, c) + \mathcal{L}_{\mathrm{BEL}}(b, c) \text{,}
\end{equation}
where $c\in \mathcal{C}$ is the ground truth label and thus $\mathbb{P} = [0,1]^{\vert \mathcal{C} \vert}$. $\mathcal{L}_{\mathrm{CE}}$ is the categorical cross entropy loss and $\mathcal{L}_{\mathrm{BEL}}$ is our novel bag embedding loss (see Sec.~\ref{BEL}).

\subsection{Preprocessing: patch and feature extraction}
Similar to most WSI classification approaches, as a preprocessing step we patch the WSI and extract features for every patch having sufficient tissue coverage. To estimate the tissue coverage, we apply Otsu's thresholding~\cite{otsu} to obtain the tissue mask $M = \text{Otsu}(S)$ where $S$ is the saturation channel of WSI $I$ in HSV colour space. All patches with less than 50\% tissue coverage are discarded and remaining patches $\mathrm{P} = \{p_1,\ldots, p_N\}$ are used for feature extraction. \par
In this preprocessing step we use a pretrained DenseNet-121~\cite{densenet}, to transform the patches into low dimensional feature vectors. Thus, we intercept the output $h \in \mathbb{R}^{K\times K \times L}$ of the fourth dense block and apply 2D global average pooling to obtain a feature vector $x\in \mathbb{R}^{L}$. Thus for each patch $p_i \in \mathrm{P}$ we compute the corresponding feature vector $x_i$ and group them into a set $X=\{x_1,\ldots,x_N\}$, which is the input to the downstream MIL model.

\subsection{Multiple instance learning using TransMIL}\label{mil:TMIL}
Contrary to typical supervised learning methods where the goal is to predict a target variable for every given input, multiple instance learning (MIL) aims to predict a single target variable $c\in \mathcal{C}$ for a set $X$ of $N$ instances or instance features, called a bag. Using an embedding based approach we perform classification by first computing a bag-level representation using the TransMIL~\cite{transmil} encoder $f_{\mathrm{ENC}}$ and thereafter applying a bag classifier $f_{\mathrm{BC}}$ to obtain class probabilities. The former is obtained by prefixing a learnable embedding $x_{cls}$ to the sequence of input instances, called class token~\cite{bert}, whose state $z_{cls}$ at the output of the encoder is denoted as the bag-level representation. \par
To be precise, the input bag $X$ of $f_{\mathrm{TMIL}}$ (eq.~\ref{eq:TMIL}) is first processed by the encoder $f_{\mathrm{ENC}}$, whose output is given by 
\begin{equation}
    z_{cls}, \{z_1,\ldots,z_N\}, \alpha = f_{\mathrm{ENC}}(\{x_1,\ldots,x_N\}; \phi, x_{cls})
\end{equation}
where $\phi, x_{cls}$ are the learnable parameters and the vector $\alpha \in [0,1]^N$ contains the attention scores of the classification token. Attention scores are obtained by extracting the first row of the attention matrix after the softmax in the last self-attention layer. Afterwards the bag classifier computes the class probabilities $p=f_{\mathrm{BC}}(z_{cls};\psi)$, where $\psi$ are its parameters. Additionally, we apply mean pooling to all output instances of the encoder to compute the bag embedding $b=\frac{1}{N}\sum\nolimits_{i=1}^{N}{z_i}$. Concatenating all of the above operations then yields $f_{\mathrm{TMIL}}$ (eq.~\ref{eq:TMIL}) with $p,b,\alpha$ as the outputs and $\theta=\{\phi,x_{cls},\psi\}$ as the learnable parameters.

\subsection{Bag embedding loss}\label{BEL}
Bag Embedding Loss (BEL) is a similarity-based loss that encourages a transformer model to learn discriminative bag-level representations. This is achieved by minimizing the distance between embeddings of the same class, while increasing the distance between embeddings of different classes. It is calculated based on the current bag embedding $b$, the corresponding ground truth label $c \in \mathcal{C}$ and the class-wise set of old bag embeddings $B=\{b_{c_1,old},\ldots , b_{c_{\vert \mathcal{C} \vert},old}\}$. Writing the new input bag embedding of a particular class $c$ as $b_{c,new}$ and having obtained the set $B$ completely, we compute the bag embedding loss as
\begin{align}\label{BEL:calc}
    \begin{split}
        \mathrm{BEL}(b_{c,new}, B) = \frac{1}{2} \left(1 - \mathcal{S}\left(b_{c,new},b_{c,old}\right)\right) \\
        +\frac{1}{2(\vert \mathcal{C} \vert -1)} \sum\limits_{\substack{c^\prime \in \mathcal{C} \\ c^\prime \neq c }} {\max\left( 0, \mathcal{S}\left(b_{c,new},b_{c^\prime,old}\right)- m\right) }
    \end{split}
\end{align}
where $m \in [0,1)$ is the margin and $\mathcal{S}(b_1, b_2) = \frac{b_1 \cdot b_2^T}{\max\left(\Vert b_1 \Vert \cdot \Vert b_2 \Vert, \epsilon\right)}$, $b_1,b_2\in \mathbb{R}^{L}$ is the cosine similarity, with $\epsilon>0$ a scalar for numerical stability. Moreover, the effect of BEL depends strongly on the quality of the class representations of the elements of $B$. Therefore we update $B$ by calculating the exponentially moving average of $b_{c,new}$ with the corresponding $b_{c,old}$ as $b_{c,old} \gets \lambda \cdot b_{c,old} + (1-\lambda) \cdot  b_{c,new}$,
where $\lambda \in [0,1)$ is the update ratio. At the start of training $B$ is initialized as an empty set and for each iteration $b_{c,new}$ is stored in $B$ until the set is complete, i.e. $\vert B \vert = \vert \mathcal{C} \vert$. Algorithm~\ref{alg:BEL} shows pseudocode for computing BEL in a training loop.
\SetKwFor{For}{for}{}{endfor}
\SetKwInput{Param}{Parameters} 
\SetKwInput{Init}{Initialize}
\begin{algorithm}
\caption{Computing BEL in a training loop}\label{alg:BEL}
\Param{Margin $m\in [0,1)$ and $\lambda \in [0,1)$}
\Init{Empty set $B=\{\}$}
\KwIn{$b_{c,new}$: Bag embedding $b$ and its label $c\in C$}
\For{$X, c$ in dataloader}{
    $p, b$, \_ = $model(X)$\\
    $celoss= \mathcal{L}_{\mathrm{CE}}(p,c)$\\
    $b_{c,new} \gets b, c$\\
    $beloss = 0$\\
    \eIf{\textbf{not} $\vert B \vert = \vert \mathcal{C} \vert$}{
        $b_{c,old} \gets b_{c,new}$\\
        $B \gets B \cup \{b_{c,old}\}$
        }{
        $beloss = \mathrm{BEL}\left(b_{c,new}, B\right)$\\
        $b_{c,old} \gets \lambda \cdot b_{c,old} + (1-\lambda) \cdot  b_{c,new}$\\
        $B \gets B \cup \{b_{c,old}\}$
        }
    $total\;loss = celoss + beloss$
    }
\end{algorithm}

\section{Experiments and results}
\begin{figure*}[h!]
    \centering
    \includegraphics[width=0.81\textwidth, keepaspectratio, page=1, trim=0cm 0.29cm 0cm 0cm,clip]{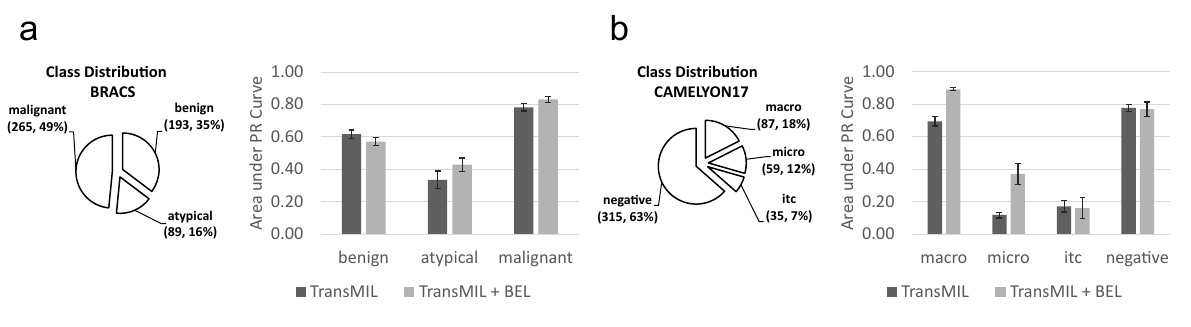}
    \caption{Class distribution and area under precision recall (PR) curve for the experiments on (a)~BRACS and (b)~CAMELYON17. For each experiment and every class we report the mean and standard deviation of a 5-fold cross validation.}
    \label{res_fig}
\end{figure*}
\subsection{Datasets}
We compare TransMIL trained with and without BEL and with our baseline DSMIL and CLAM on two publicly available histopathological datasets: \\

\noindent\textbf{BRACS.} The BReast Cancer Subtyping (BRACS) dataset consists of $547$ H\&E stained WSIs obtained from $189$ patients, with breast carcinoma~\cite{bracs}. The WSIs were selected from the archives of the Department of Pathology at National Cancer Institute- IRCCS-Fondazione Pascale, Naples, Italy. They were scanned with an Aperio AT2 scanner at $0.25$ $\mu$m/pixel for $40\times$ magnification. There three main disease classes are labelled: 'benign', 'atypical' and 'malignant'. We use the train, validate and test sets as provided by the dataset developers.\linebreak

\noindent\textbf{CAMELYON17.} The CAMELYON17 dataset contains 1000 WSIs with 5 slides per patient and with 500 slides in train and test sets each~\cite{camelyon17}. In contrast to other recent works~\cite{DSMIL,CLAM,transmil}, where slides have been classified in a binary fashion, we perform a classification into 3 metastasis types: 'macro', 'micro', and 'itc', as well as slides containing no metastasis designated as 'negative'. The respective labels are only available in CAMELYON17. Since no slide-level annotations are available for the test set, we focus exclusively on the training set, where we perform patient-wise stratified splits.


\subsection{Implementation details}
The proposed method consists of four components (Fig.~\ref{con_fig}): Patch extraction, feature extraction, TransMIL and BEL:

\noindent (i) For patch extraction WSIs are processed at $40\times$ magnification and divided into patches of size $512\times 512$. The maximum number of patches extracted per WSI was $24635$ for BRACS and $21053$ for CAMELYON17.

\noindent (ii) We perform feature extraction using KimiaNet~\cite{kimianet}, a finetuned version of the DenseNet-121 architecture~\cite{densenet}, which was specifically trained on the TCGA dataset for the tasks of cancer subtype classification and tumor vs. normal tissue classification. The latent dimension of our features is $L=1024$.

\noindent (iii) As suggested by the authors, TransMIL is trained with RAdam~\cite{radam} which is wrapped by a lookahead optimizer~\cite{lookahead}. We use a learning rate of $0.00002$, a weight decay of $0.00005$ and an attention dropout probability of $0.2$. The elements of the output sequence have $512$ dimensions. For training DSMIL~\cite{DSMIL} and CLAM~\cite{CLAM} we use the optimizer and hyper parameters from the author's repository. All models are trained for $100$ epochs on each fold. 

 \noindent (iv) BEL has two hyper parameters: The margin $m$ and the update ratio $\lambda$. We observed that a margin $m \in [0.1, 0.25]$ performs best and we used $m=0.25$ in our experiments. For the update ratio we used $\lambda=0.996$.
\begin{table}[h]
    \caption{Comparison of TransMIL when trained with and without BEL on two datasets using DSMIL, CLAM Single Branch (SB) and CLAM Multi Branch (MB) as baselines. For each experiment we report the mean and standard deviation for accuracy, F1-score and area under ROC (AUROC) curve from a 5-fold cross validation.}
    
    \begin{adjustbox}{width=\columnwidth,center}
    \begin{tabular}{p{2cm}|l||c|c|c}
        Dataset                                                                              & Method & Accuracy & F1 score &  AUROC  \\ 
        \hline \hline
        \multirow{5}{*}{\begin{tabular}[c]{@{}l@{}}BRACS\end{tabular}}                       & DSMIL & $0.56\pm 0.01$ & $0.50\pm 0.05$ & $0.74\pm 0.02$  \\ 
        \cline{2-5}
                                                                                             & CLAM SB & $0.53\pm 0.01$ & $0.48\pm 0.03$ & $0.70\pm 0.03$ \\
        \cline{2-5}
                                                                                             & CLAM MB & $0.54\pm 0.04$ & $0.51\pm 0.06$ & $0.72\pm 0.01$ \\
        \cline{2-5}
                                                                                             & TransMIL & $0.54\pm 0.04$ & $0.49\pm 0.05$ & $0.72\pm 0.01$ \\ 
        \cline{2-5}
                                                                                             & TransMIL + BEL & $\mathbf{0.60\pm 0.02}$  & $\mathbf{0.57\pm 0.01}$ & $\mathbf{0.76\pm 0.01}$ \\ 
        \hline \hline
        \multirow{5}{*}{\begin{tabular}[c]{@{}l@{}}CAMELYON\end{tabular}}                    & DSMIL  & $0.68\pm 0.04$ & $0.40\pm 0.02$ & $0.72\pm 0.02$  \\ 
        \cline{2-5}
                                                                                             & CLAM SB & $0.71\pm 0.02$ & $0.46\pm 0.04$ & $0.73\pm 0.01$ \\
        \cline{2-5}
                                                                                             & CLAM MB & $0.71\pm 0.02$ & $0.44\pm 0.03$ & $0.76\pm 0.03$ \\
        \cline{2-5}
                                                                                             & TransMIL & $0.68\pm 0.01$ & $0.39\pm 0.04$ & $0.68\pm 0.02$  \\ 
        \cline{2-5}
                                                                                             & TransMIL + BEL & $\mathbf{0.73\pm 0.01}$  & $\mathbf{0.48\pm 0.02}$ & $\mathbf{0.75\pm 0.04}$ \\ 
    \end{tabular}
    \end{adjustbox}
    \label{res_tab}
\end{table}

\subsection{Results}
For training our models we split our data into a training and a hold out test set using the ratio 5:1 and perform a 5-fold cross validation on the training set. We then take the model with the highest validation accuracy from each fold and evaluate it on the hold out test set. For each experiment, we report accuracy, macro F1 score and area under the ROC (AUROC) curve in Table~\ref{res_tab}. In both datasets, thanks to BEL, TransMIL is able to outperform all baseline models, with a significant performance improvement compared to TransMIL trained without BEL.
To get an insight into the class-wise performance, we compare TransMIL with and without BEL. Figure~\ref{res_fig} shows the area under the precision-recall curve for both datasets and all classes. We found that the improvements mainly resides in disease classes which are not strongly represented, expect for 'itc' in the CAMELYON17 dataset.  




\section{Discussion \& Conclusion}
Our proposed Bag Embedding Loss (BEL) successfully improves TransMIL's performance. Our findings suggest that adding a supervised signal helps to overcome the challenges posed by weak annotations and large bag sizes. We assume that the improvement in performance is largely due to the distance maximization in the second term loss term (eq.~\ref{alg:BEL}). Our approach holds promise for enhancing the classification of histopathology whole slide images for clinical applications.

\section{Compliance with Ethical Standards}
This research study was conducted retrospectively using human subject data made available in open access. Ethical approval was not required as confirmed by the license attached with the open access data.

\section{Acknowledgements}
C.M. has received funding from the European Research Council (ERC) under the European Union’s Horizon 2020 research and innovation programme (Grant agreement No. 866411)

\bibliographystyle{IEEEbib}
\bibliography{refs}

\begin{thebibliography}{10}

\bibitem{pooling1}
Xinggang Wang, Yongluan Yan, Peng Tang, Xiang Bai, and Wenyu Liu,
\newblock ``Revisiting multiple instance neural networks,''
\newblock {\em Pattern Recognition}, vol. 74, pp. 15--24, Feb 2018.

\bibitem{pooling2}
Fahdi Kanavati, Gouji Toyokawa, Seiya Momosaki, Michael Rambeau, Yuka Kozuma,
  Fumihiro Shoji, Koji Yamazaki, Sadanori Takeo, Osamu Iizuka, and Masayuki
  Tsuneki,
\newblock ``Weakly-supervised learning for lung carcinoma classification using
  deep learning,''
\newblock {\em Scientific Reports}, vol. 10, no. 1, pp. 9297, Jun 2020.

\bibitem{ABMIL}
Maximilian Ilse, Jakub~M. Tomczak, and Max Welling,
\newblock ``Attention-based deep multiple instance learning,'' 2018.

\bibitem{CLAM}
Ming~Y. Lu, Drew F.~K. Williamson, Tiffany~Y. Chen, Richard~J. Chen, Matteo
  Barbieri, and Faisal Mahmood,
\newblock ``Data efficient and weakly supervised computational pathology on
  whole slide images,'' 2020.

\bibitem{DSMIL}
Bin Li, Yin Li, and Kevin~W. Eliceiri,
\newblock ``Dual-stream multiple instance learning network for whole slide
  image classification with self-supervised contrastive learning,''
\newblock in {\em Proceedings of the IEEE/CVF Conference on Computer Vision and
  Pattern Recognition (CVPR)}, June 2021, pp. 14318--14328.

\bibitem{ADMIL}
Andriy Myronenko, Ziyue Xu, Dong Yang, Holger~R. Roth, and Daguang Xu,
\newblock ``Accounting for dependencies in deep learning based multiple
  instance learning for whole slide imaging,''
\newblock in {\em Medical Image Computing and Computer Assisted Intervention --
  MICCAI 2021}, Cham, 2021, pp. 329--338, Springer International Publishing.

\bibitem{DTMIL}
Hang Li, Fan Yang, Yu~Zhao, Xiaohan Xing, Jun Zhang, Mingxuan Gao, Junzhou
  Huang, Liansheng Wang, and Jianhua Yao,
\newblock ``{DT-MIL: Deformable Transformer for Multi-instance Learning on
  Histopathological Image},''
\newblock in {\em Medical Image Computing and Computer Assisted Intervention --
  MICCAI 2021}, Cham, 2021, pp. 206--216, Springer International Publishing.

\bibitem{transmil}
Zhuchen Shao, Hao Bian, Yang Chen, Yifeng Wang, Jian Zhang, Xiangyang Ji, and
  Yongbing Zhang,
\newblock ``Transmil: Transformer based correlated multiple instance learning
  for whole slide image classification,'' 2021.

\bibitem{nystroemformer}
Yunyang Xiong, Zhanpeng Zeng, Rudrasis Chakraborty, Mingxing Tan, Glenn Fung,
  Yin Li, and Vikas Singh,
\newblock ``Nystr{\"o}mformer: A nystr{\"o}m-based algorithm for approximating
  self-attention,''
\newblock 2021.

\bibitem{densenet}
Gao Huang, Zhuang Liu, Laurens van~der Maaten, and Kilian~Q. Weinberger,
\newblock ``Densely connected convolutional networks,'' 2018.

\bibitem{otsu}
Nobuyuki Otsu,
\newblock ``A threshold selection method from gray-level histograms,''
\newblock {\em IEEE Transactions on Systems, Man, and Cybernetics}, vol. 9, no.
  1, pp. 62--66, 1979.

\bibitem{bert}
Jacob Devlin, Ming-Wei Chang, Kenton Lee, and Kristina Toutanova,
\newblock ``Bert: Pre-training of deep bidirectional transformers for language
  understanding,'' 2019.

\bibitem{bracs}
Nadia Brancati, Anna~Maria Anniciello, Pushpak Pati, Daniel Riccio, Giosuè
  Scognamiglio, Guillaume Jaume, Giuseppe~De Pietro, Maurizio~Di Bonito,
  Antonio Foncubierta, Gerardo Botti, Maria Gabrani, Florinda Feroce, and Maria
  Frucci,
\newblock ``Bracs: A dataset for breast carcinoma subtyping in h\&e histology
  images,'' 2021.

\bibitem{camelyon17}
P\'eter B\'andi, Oscar Geessink, Quirine Manson, Marcory Van~Dijk, Maschenka
  Balkenhol, Meyke Hermsen, Babak Ehteshami~Bejnordi, and et~al.,
\newblock ``From detection of individual metastases to classification of lymph
  node status at the patient level: The camelyon17 challenge,''
\newblock {\em IEEE Transactions on Medical Imaging}, vol. 38, no. 2, pp.
  550--560, 2019.

\bibitem{kimianet}
Abtin Riasatian, Morteza Babaie, Danial Maleki, Shivam Kalra, Mojtaba Valipour,
  Sobhan Hemati, Manit Zaveri, Amir Safarpoor, and Sobhan Shafiei,
\newblock ``Fine-tuning and training of densenet for histopathology image
  representation using tcga diagnostic slides,''
\newblock {\em Medical Image Analysis}, vol. 70, pp. 102032, 2021.

\bibitem{radam}
Liyuan Liu, Haoming Jiang, Pengcheng He, Weizhu Chen, Xiaodong Liu, Jianfeng
  Gao, and Jiawei Han,
\newblock ``On the variance of the adaptive learning rate and beyond,'' 2021.

\bibitem{lookahead}
Michael~R. Zhang, James Lucas, Geoffrey Hinton, and Jimmy Ba,
\newblock ``Lookahead optimizer: k steps forward, 1 step back,'' 2019.

\end{thebibliography}

\end{document}